\documentclass[10pt,twocolumn,letterpaper]{article}

\usepackage{cvpr}              

\usepackage{graphicx}
\usepackage{amsmath}
\usepackage{amssymb}
\usepackage{booktabs}

\usepackage{epsfig}
\usepackage{graphics}
\usepackage{color}
\usepackage{mathtools}
\usepackage{multirow}
\usepackage{tabularx}
\usepackage{amsfonts}
\usepackage{wrapfig,lipsum,booktabs}
\usepackage{xcolor}
\usepackage{floatflt}
\usepackage{caption}
\usepackage{subfloat}
\usepackage[ruled, lined, linesnumbered, commentsnumbered, longend]{algorithm2e}
\usepackage{algpseudocode}

\usepackage[cal=cm]{mathalfa} 
\usepackage{bbm, dsfont}

\colorlet{dark-blue}{blue!70!black}
\colorlet{dark-green}{green!50!black}
\colorlet{dark-red}{red!80!black}
\usepackage[normalem]{ulem}

\usepackage[pagebackref,breaklinks,colorlinks]{hyperref}

\usepackage[capitalize]{cleveref}
\crefname{section}{Sec.}{Secs.}
\Crefname{section}{Section}{Sections}
\Crefname{table}{Table}{Tables}
\crefname{table}{Tab.}{Tabs.}

\def\ie{\textit{i.e.}}

\usepackage{caption}
\usepackage{enumitem}
\setitemize{noitemsep,topsep=0pt,parsep=0pt,partopsep=0pt}

\def\cvprPaperID{1326} 
\def\confName{CVPR}
\def\confYear{2022}

\begin{document}

\title{Bi-Mix: Bidirectional Mixing for \\ Domain Adaptive Nighttime Semantic Segmentation}

\author{Guanglei Yang$^{1,2}$ \quad Zhun Zhong$^{2}$ \quad  Hao Tang$^3$ \quad Mingli Ding$^1$ \quad Nicu Sebe$^2$ \quad Elisa Ricci$^{2,4}$\\
	$^1$Harbin Institute of Technology, China  \quad $^2$DISI, University of Trento, Italy \\
	$^3$Computer Vision Lab, ETH Zurich, Switzerland  \quad $^4$Fondazione Bruno Kessler, Italy
}

\maketitle

\begin{abstract}

In autonomous driving, learning a segmentation model that can adapt to various environmental conditions is crucial. In particular, copying with severe illumination changes is an impelling need, as models trained on daylight data will perform poorly at nighttime. In this paper, we study the problem of Domain Adaptive Nighttime Semantic Segmentation (DANSS), which aims to learn a discriminative nighttime model with a labeled daytime dataset and an unlabeled dataset, including coarsely aligned day-night image pairs. To this end, we propose a novel Bidirectional Mixing (Bi-Mix) framework for DANSS, which can contribute to both image translation and segmentation adaptation processes. Specifically, 
in the image translation stage, Bi-Mix leverages the knowledge of day-night image pairs to improve the quality of nighttime image relighting. On the other hand, in the segmentation adaptation stage, Bi-Mix effectively bridges the distribution gap between day and night domains for adapting the model to the night domain. In both processes, Bi-Mix simply operates by mixing two samples without extra hyper-parameters, thus it is easy to implement.
Extensive experiments on Dark Zurich and Nighttime Driving datasets demonstrate the advantage of the proposed Bi-Mix and show that our approach obtains state-of-the-art performance in DANSS.
Our code is available at \url{https://github.com/ygjwd12345/BiMix}.

\end{abstract}
\section{Introduction}

\begin{figure}[t]
    \centering
    \includegraphics[width=1\linewidth]{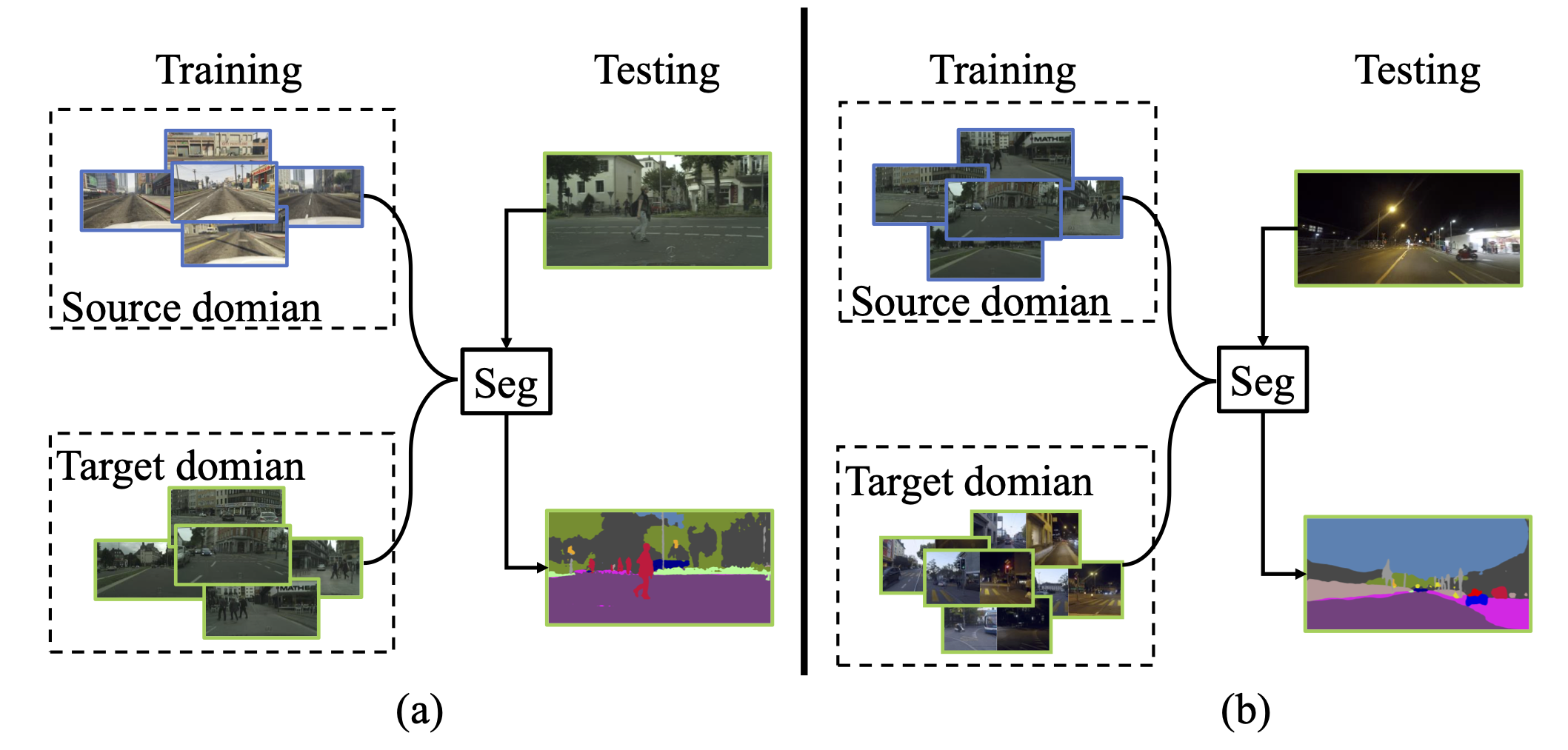}
    \caption{Comparison of (a) traditional domain adaptive semantic segmentation and (b) domain adaptive nighttime semantic segmentation. ``Seg'' denotes the semantic segmentation network.
    }
    \label{fig:teaser}
\end{figure}

In the past few years, domain adaptive semantic segmentation~\cite{yang2020fda,zhang2021prototypical,wang2021domain,gong2021cluster}, which aims to transfer the knowledge from a labeled source domain to an unlabeled target domain, has been widely studied. Most existing methods~\cite{gong2019dlow,vu2019advent,li2019bidirectional,zhang2021prototypical,wang2021domain} are typically tested in situations of limited \emph{domain shift}, \ie,~when both source and target images are collected in favourable weather and illumination conditions.  
For instance, the widely used CityScapes~\cite{cordts2016cityscapes} dataset only comprises images captured during sunny daytime. However, in real-world autonomous driving scenes, data may be collected under several unfavorable conditions, such as rain, snow, fog, or nighttime. As a consequence, a model trained on existing datasets will produce poor performance on images gathered under the above tricky settings. Therefore, it is important to devise strategies in order to learn a robust segmentation model that can generalize well to unfavorable conditions. In this paper, we study this problem and in particular we address the Domain Adaptive Nighttime Semantic Segmentation (DANSS) task originally introduced in \cite{dai2018dark}. 
The difference between traditional domain adaptive semantic segmentation and DANSS is shown in Figure~\ref{fig:teaser}.
The goal of DANSS is to learn a robust model for nighttime with a fully labeled source dataset of daytime images and an unlabeled target dataset that includes coarsely aligned day-night image pairs. Note that DANSS can be regarded as a multi-target domain adaptation problem. This is because (1) the two datasets are from different domains, and (2) the target dataset contains two sub-domains, \textit{i.e.}, daytime domain and nighttime domain.

To solve the problem of DANSS, previous methods~\cite{dai2018dark,sakaridis2018semantic,sakaridis2019guided} utilize an intermediate twilight domain to transfer the model from daytime domain to twilight domain gradually and then from twilight domain to nighttime domain. In such solutions, they adopt a translation-then-segmentation strategy with multiple learning stages, which requires learning separate image translation models and is very time-consuming. In addition, the current learning stage is affected by previous stages' results, so the accumulated training errors may hurt the model optimization. Recently, Wu \textit{et al.} proposed a one-stage approach, called DANNet~\cite{wu2021dannet}, which integrates image translation and semantic segmentation tasks into a unified framework and does not require an extra twilight domain. Compared to previous multi-stage methods, DANNet is more efficient and can produce competitive results on the Dark Zurich dataset. Despite its success, DANNet neglects three important factors in DANSS. First, the information between coarsely aligned day-night image pairs is ignored during image translation. Second, the segmentation model greatly benefits from the image translation model while the segmentation model has little guidance from the translation model.
Third, the issue of class imbalance in nighttime images is ignored.

{To this end, in this paper we propose a novel Bidirectional Mixing (Bi-Mix) framework for DANSS, which leverages the sample mixing technique to enhance the quality of image translation with day-night jointly paired information and overcome imbalanced class problems during adaptation. Similar to DANNet~\cite{wu2021dannet}, our Bi-Mix is an one-stage framework, which contains both an image translation model and a semantic segmentation model. However, it is a bidirectional learning framework, involving two directions, \ie,  ``translation-to-segmentation (Trans2Seg)''  and  ``segmentation-to-translation (Seg2Trans)''. Specifically, in the \textit{Trans2Seg direction}, we first relight the images by the image translation model and then predict the segmentation results of them by the segmentation model. In this direction, the image translation model aims to revive the visual contents of night images, with the purpose of facilitating the downstream segmentation process. The segmentation model is trained with a segmentation loss and an adversarial loss. In addition, we mix pixels of randomly selected classes from the source sample with nighttime sample, enabling us to train the nighttime samples in a class-balanced manner. In the \textit{Seg2Trans direction}, we use a similar mixing strategy as in Trans2Seg direction, where the difference is that we mix the daytime sample with the nighttime sample with the guidance of the predicted segmentation results. As such, the mixed sample not only contains the nighttime style but also is more consistent with the daytime sample in terms of content. It exploits the paired information between coarsely aligned day-night image pairs and enables us to use a consistency loss to improve the image translation model further. In our Bi-Mix, the learning in the two directions benefit from each other: the translation model generates favorable nighttime samples for learning a more robust segmentation model, while the segmentation model provides accurate paired samples for training more effectively the image translation model. 

\noindent\textbf{Contributions.} The contributions of this paper are:
\begin{itemize}
\vspace{.05in}
\item We introduce a Bidirectional Mixing (Bi-Mix) framework for DANSS, which bridges and promotes sinergy between the translation-adaptation and the segmentation-adaptation processes.
\vspace{.05in}
\item We propose a novel sample mixing strategy, which enables us to leverage the information between coarsely aligned day-night image pairs to improve the translation adaptation model.
\vspace{.05in}
\item Extensive experiments on two challenging unsupervised nighttime semantic segmentation benchmarks demonstrate the effectiveness of our Bi-Mix framework, which outperforms the state-of-the-art methods. 

\end{itemize}
}

\section{Related Work}
\noindent {Nighttime Semantic Segmentation} is critical in safety autonomous driving. A segmentation model trained on daytime data produces poor performance on nighttime images due to the large domain shift. In addition, annotating nighttime data is extremely difficult. Considering these factors, Dai \emph{et al.}~\cite{dai2018dark} introduced the task of Domain Adaptive Nighttime Semantic Segmentation (DANSS) and proposed a two-step adaptation approach with the help of an intermediate twilight domain to tackle this task. Sakaridis \emph{et al.}~\cite{sakaridis2019guided} used an image translation model to transfer the style of nighttime images to that of the daytime images. Latter, Sakaridis \emph{et al.}~\cite{SDV20} extended ~\cite{sakaridis2019guided} by leveraging geometry information to refine the semantic predictions. Recently, Xu \emph{et al.}~\cite{Xu2021CDAda} proposed a curriculum domain adaptation method to transfer the semantic knowledge from daytime to nighttime smoothly. All these methods require training the model on multiple stages. To reduce the training cost and avoid the issue of accumulated training error, Wu \emph{et al.}~\cite{wu2021dannet} introduced a one-stage adaptation framework, called DANNet~\cite{wu2021dannet}. DANNet jointly trains the image translation model and the semantic segmentation model in an one-stage fashion. This is more efficient and can produce comparable results with multi-stage-based methods. However, in DANNet, the information among paired daytime and nighttime images is largely ignored. 
Differently, in this work, we propose a novel Bi-Mix framework to leverage this information and improve the segmentation outputs. 

\begin{figure}[!t]
    \centering
    \includegraphics[width=1\linewidth]{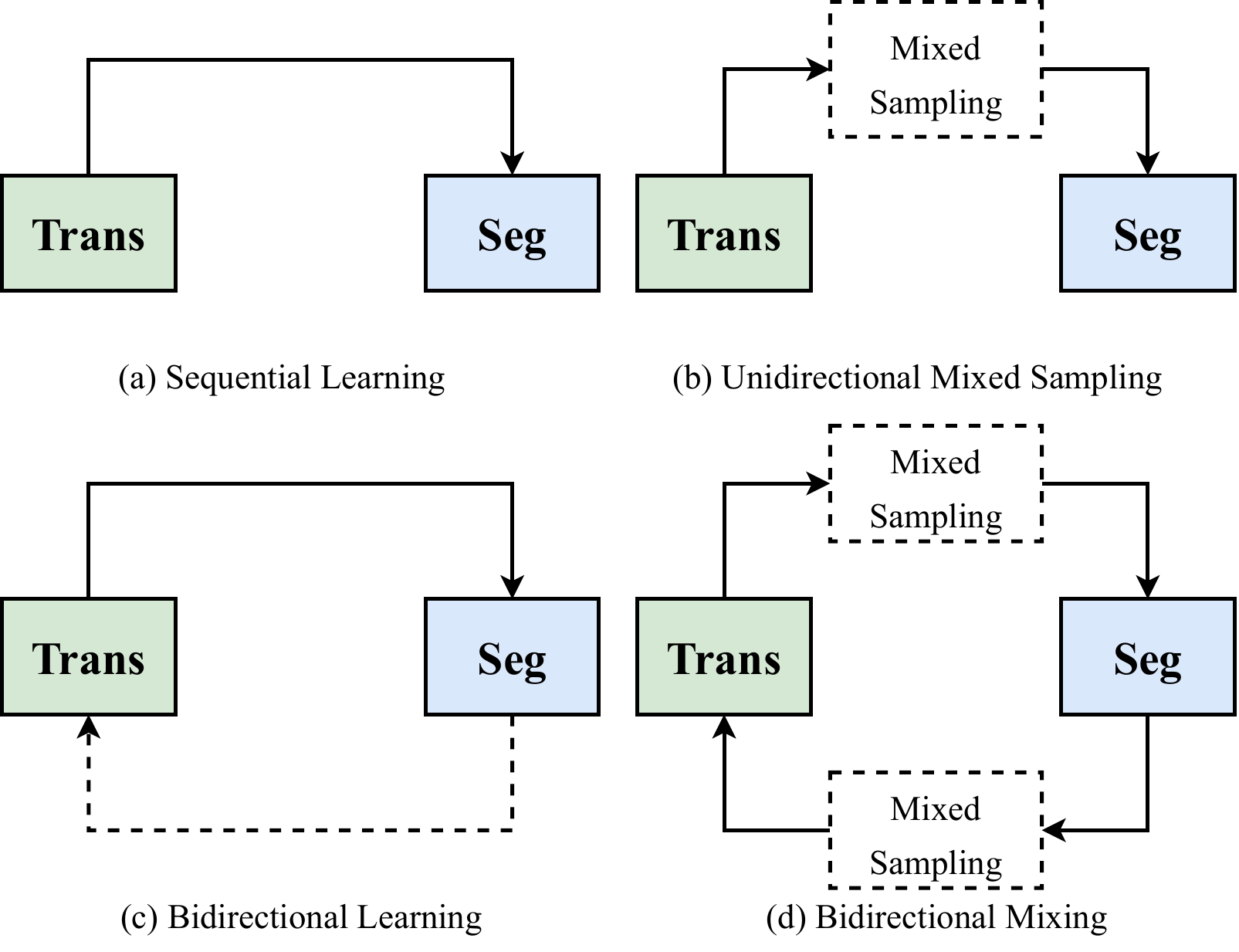}
    \caption{Comparison of different learning strategies.}
    \label{fig:bms}
\end{figure}

\section{Methodology}

\subsection{Problem Definition} 
In Domain Adaptive Nighttime Semantic Segmentation we are given a source domain $\mathcal{S}$ where a dataset $S=\{I^i_s,Y_s^i\}_{i=1}^{N_s}$ is drawn and segmentation labels $Y_s^i$ are available for images $I^i_s$ and two target domains $\mathcal{T}_d$ and $\mathcal{T}_n$, whose associated unlabeled datasets are denoted as $T_d=\{I^i_d\}_{i=1}^{N_d}$ and $T_n=\{I^i_n\}_{i=1}^{N_n}$ respectively. The two target domains correspond to daytime and nighttime images, respectively, and are coarsely paired in terms of locations. 
The goal of DANSS is to learn a model using as training data images of all the three domains, which can perform well on the nighttime test set.

\subsection{Bidirectional Learning vs Bidirectional Mixing}
Our work is inspired by Bidirectional Learning (BL).
BL, as opposed to sequential learning, has been initially proposed to learn a language translation model~\cite{he2016dual,niu2018bi}. Later, it has been applied to several visual tasks, such as image generation~\cite{pontes2019bidirectional,russo2018source} and domain adaptation~\cite{li2019bidirectional}. Figure~\ref{fig:bms} compares sequential learning and BL. Here we instantiate them in the case of a domain adaptation framework and consider two models: an image translation model and a semantic segmentation model.
As shown in Figure~\ref{fig:bms}(a) and Figure~\ref{fig:bms}(c), the translation model and the segmentation model are mutually influenced in BL, while the processing flow only goes from the translation model to the segmentation model in sequential learning. By injecting the sample mixing technique~\cite{olsson2021classmix,tranheden2021dacs}, the sequential learning strategy can be extended to sequential mixing learning (Figure~\ref{fig:bms}(b)).
In this work we propose a novel approach, \textit{i.e.}, Bidirectional Mixing Learning (Figure~\ref{fig:bms}(d)). Unlike BL, which uses the segmentation results to improve the translation model via back-propagation, we additionally use the segmentation results as guidance when generating new samples. The image generation (mixing) process is detached from the segmentation model, in which the generated samples are directly used to train the translation model in a new direction. The proposed approach is described in details in the following.

\begin{figure}[t]
    \centering
    \includegraphics[width=1\linewidth]{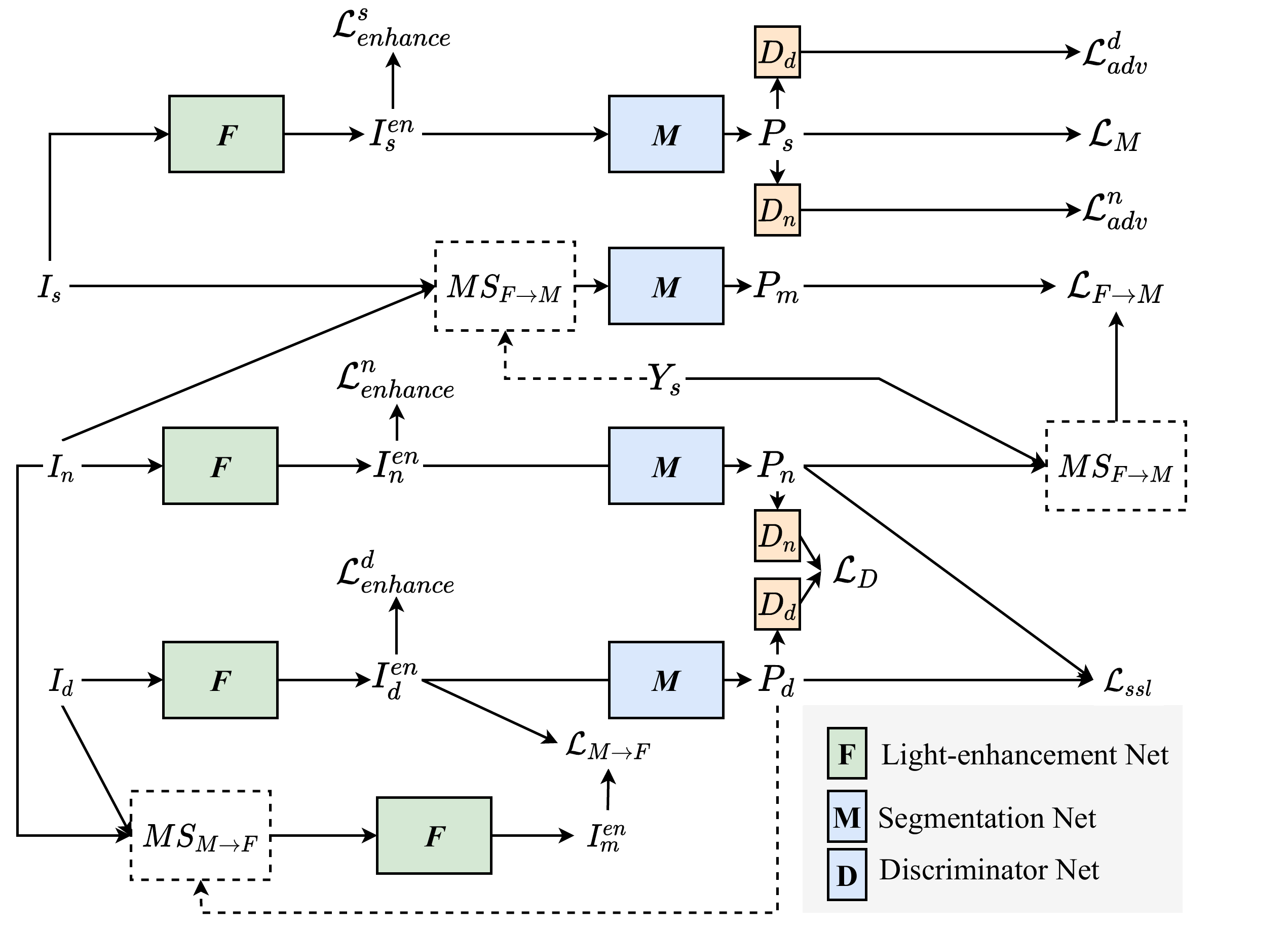}
    \caption{Overview of the proposed method. The model includes a light-enhancement or translation network $F$, a segmentation network $M$, and two discriminator networks $D_n$ and $D_d$. $MS_{F\to M}$ is indicates mixing $S$ and $T_n$ while $MS_{M\to F}$ mixing $T_d$ and $T_n$. $MS$ means mixed sampling.
    }
    \label{fig:overview}
\end{figure}

\subsection{Framework Overview} 
An overview of the proposed method is shown in Figure~\ref{fig:overview}. Our framework contains a translation or light-enhancement network $F$, a segmentation network $M$, and two discriminator networks $D_n$ and $D_d$. Three images, sampled from $S$, $T_n$, and $T_d$, 
respectively, are provided as input. We first feed the inputs into the network $F$ to obtain light-enhanced counterparts. We then calculate the enhancement losses on them for optimizing $F$. 
{The enhanced images, $I_s^{en}$, $I_d^{en}$, and $I_n^{en}$, are then fed into the segmentation model to predict the segmentation results. A semantic loss $\mathcal{L}_M$
is computed using the source label $Y_s$ and source segmentation prediction $P_s$. Moreover, the prediction $P_d$ from the daytime image $I_d$
is used as a pseudo label to provide weak supervision for the paired prediction $P_n$ with a self-supervised loss $\mathcal{L}_{ssl}$. We also use adversarial learning to learn domain-invariant outputs, where the adversarial losses are calculated on $P_s$, $P_d$ and $P_n$.
For the Bidirectional Mixing, \textit{at the segmentation-level}, the output of mixed sampling between images of $S$ and $T_n$ is fed into the same segmentation network, which is supervised by a cross-entropy loss $\mathcal{L}_{F\to M}$. \textit{At the translation-level}, we use the prediction of a daytime image to guide the mixing of between images of $T_n$ and $T_d$. The mixed sample is fed into $F$, and a consistency loss $\mathcal{L}_{M\to F}$ is used. 
}

\begin{figure*}[t]
    \centering
    \includegraphics[width=1\linewidth]{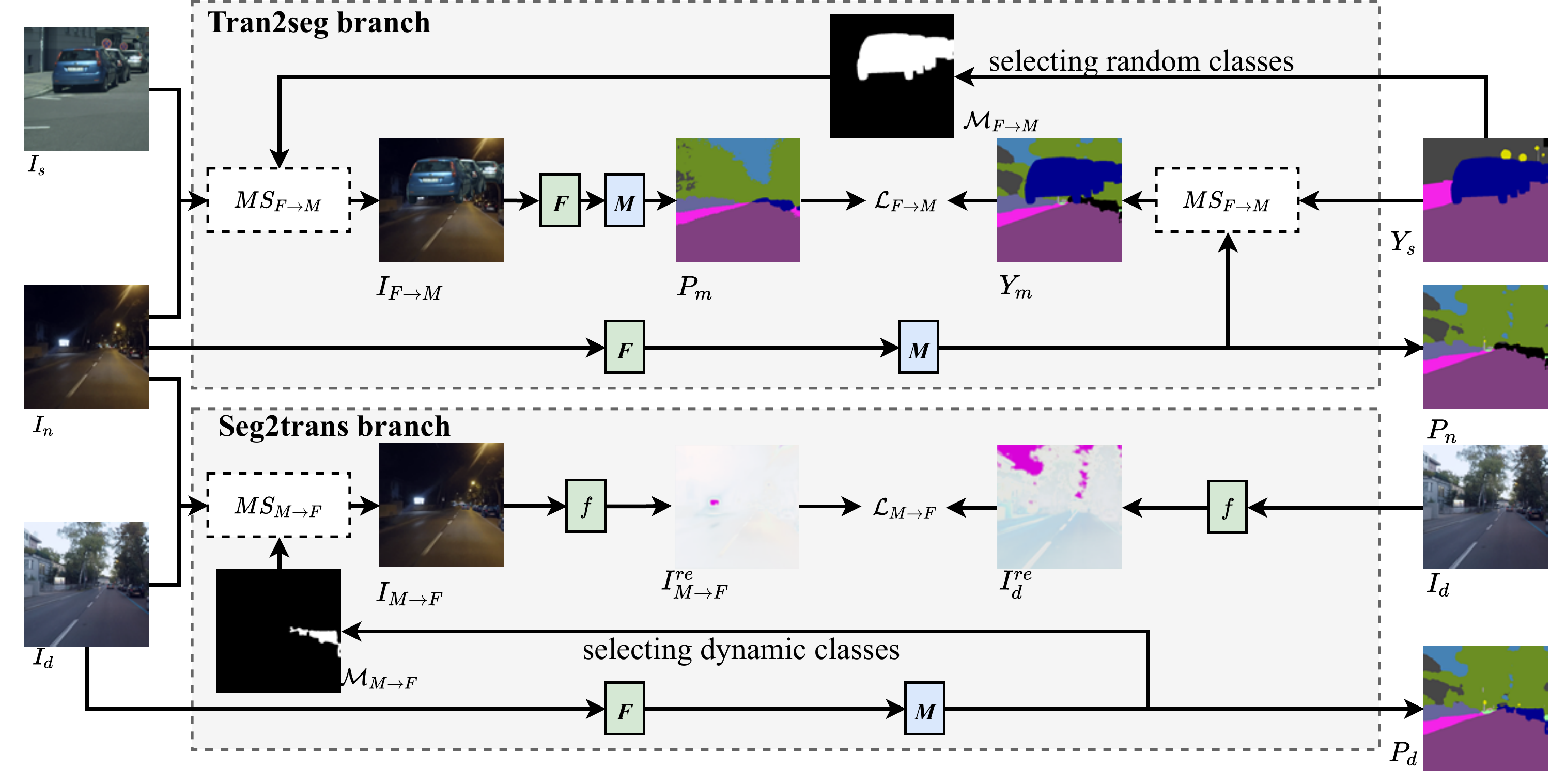}
    \caption{Overview of Bidirectional Mixing. $F$, $M$, and $f$ denote the light-enhancement network, the segmentation network and the relighting network, respectively. The relationship between $F$ and $f$ is defined in Eq.~\eqref{eq:f}.}
    \label{fig:d-bms}
\vspace{-0.4cm}    
\end{figure*}

\subsection{Bidirectional Mixing}
\label{sec:bms}

{In this paper, we propose a novel data augmentation technique, called Bidirectional Mixing (Bi-Mix), for DANSS.
Bi-Mix leverages both the labeled source and unlabeled target samples to synthesize new images, effectively bridging and promoting the translation adaptation and the segmentation adaptation process jointly.}
Figure~\ref{fig:d-bms} illustrates the process of our Bi-Mix. The whole Bi-Mix involves two unidirectional sample mixing branches, \textit{i.e.}, the Trans2Seg branch and the Seg2Trans branch. 

\noindent \textbf{Trans2Seg Branch.} In this branch, we first feed images to the light enhancement network $F$, where the outputs are then forwarded to the segmentation network $M$.
The light enhancement network is an image-to-image translation network. The output of the light enhancement network can be calculated as:

\begin{equation}
I^{en}=I^{re}+I=f(I)+I,
\label{eq:f}
\end{equation}
where $f$ is a relighting network. $I$ is the input image, which is sampled from the datasets $S$, $T_d$ or $T_n$.

Inspired by ClassMix~\cite{olsson2021classmix,yun2019cutmix}, the unidirectional mixed sampling on the Trans2Seg branch is applied on labeled source images and unlabelled night images.
Specifically, given all classes in $Y_s$ is $\mathcal{C}_s$, half of the classes among $\mathcal{C}_s$ are randomly chosen as mixing classes $\mathcal{C}_s^m$. We then generate a binary mask $\mathcal{M}_{F\to M}$ by:
\begin{equation} \label{eq:mask-1} \small
    \mathcal{M}_{F\to M}=\left \{\begin{array}{lc}
         1, & Y_s^{(p)}\in \mathcal{C}_s^m \\
         0,  & otherwise
    \end{array} \right. ,
\end{equation}
where $p$ denotes a pixel in the source image. Based on $\mathcal{M}_{F\to M}$, the mixed source image $I_{F\to M}$ is generated by:
\begin{equation}
    I_{F\to M}= \mathcal{M}_{F\to M}\cdot I_s + (1-\mathcal{M}_{F\to M})\cdot I_n,
\end{equation}

The same mixing process is also applied to the predictions $P_n$ and ground truth $Y_s$, resulting in the mixed label $Y_m$. The cross-entropy loss is used to optimize the model:
\begin{equation}
\label{eq:mix_ce}
    \mathcal{L}_{F\to M}=-\frac{1}{N\cdot C}\sum_{c\in C} \mathbbm{1}(Y_m)^c\log P_m^c,
\end{equation}
where $P_m^c$ is the $c$-th channel of the prediction $P_m$ and $\mathbbm{1}(\cdot)$ indicates the one-hot encoding operation.
In this manner, we inject the knowledge of the labeled sample into the nighttime image, enabling us (1) to be robust for noisy prediction associated to nighttime images and  (2) to optimize the model in a more class-balanced way.

We also introduce a self-supervised loss to leverage the prediction for the daytime images as the pseudo labels and refining the corresponding prediction for the nighttime images. To account for the imbalance among different categories of nighttime images, we adopt a focal loss~\cite{lin2017focal} instead of vanilla cross-entropy loss. This loss $\mathcal{L}_{ssl}$ can be expressed as:
\begin{equation}
    \mathcal{L}_{ssl}=-\frac{1}{N}||(1-p_d)^\gamma \log(p_n)||_1,
\end{equation}
where $\gamma$ is the focusing parameter (set to 1 in all our experiments) and $N$ is the number of the pixels in an image. $p_d$ and $p_n$ are the likelihood map of the predicted categories. Specifically, the predicted categories are estimated by choosing the categories that have the largest probabilities in the daytime prediction $P_d$. 

{The light-enhancement network is supervised by several losses, including the total variation loss $\mathcal{L}_{tv}$, the exposure control loss $\mathcal{L}_{exp}$, and the structural similarity loss $\mathcal{L}_{ssim}$. These three losses enable us to learn a relighting network in a self-supervised manner.

In detail, we apply $\mathcal{L}_{tv}$ to avoid noise artifacts 
influencing the semantic segmentation. Indicating with a generic input image $I$ from $S$, $T_n$ and $T_d$ and with  the corresponding output image $I^{re}$ from the relighting network, $\mathcal{L}_{tv}$ is defined by:
\begin{equation}
    \mathcal{L}_{tv}=\frac{1}{N}||(\nabla_x(I-I^{re}))^2+\nabla_y(I-I^{re}))^2||_1,
\end{equation}
where $N$ is the number of pixels in $I$ and $\nabla_x$ and $\nabla_y$ represent horizontal and vertical gradient operations.
The loss $\mathcal{L}_{exp}$ is designed to control the exposure level, which can be expressed as:
\begin{equation}
    \mathcal{L}_{exp}=\frac{1}{N_R}||\psi(I^{re})-I^{en}||_1,
\end{equation}
where $\psi$ is an average pooling function and $N_R$ represents the number of pixels in $\psi(I^{re})$.

Finally, the structural similarity loss $\mathcal{L}_{ssim}$~\cite{wang2004image} is used to prevent the image degradation caused by image translation process.  $\mathcal{L}_{ssim}$ can be formulated as:
\begin{equation}
\mathcal{L}_{ssim}=\frac{1}{2N}||1-SSIM(I,I^{re})||_1.
\end{equation}
The overall enhancement loss $\mathcal{L}_{enhance}$ is defined by
\begin{equation}
    \mathcal{L}_{enhance}=\alpha_{tv}\mathcal{L}_{tv}+ \alpha_{exp}\mathcal{L}_{exp}+ \alpha_{ssim}\mathcal{L}_{ssim},
\end{equation}
where $\alpha_{tv}$, $\alpha_{exp}$ and $\alpha_{ssim}$ are  hyperparameters.

\noindent \textbf{Seg2Trans Branch.} In the reverse direction, we aim to improve the performance of the light enhancement network by leveraging the semantic information produced by the segmentation network. Similarly to the mixed sampling strategy, two unlabelled images, $I_d$ and $I_n$, are sampled from the target datasets. Both of them are fed through the relighting network $f$, whose outputs are $I_d^{re}$ and $I_n^{re}$ .
Given all the classes in $P_d$ is $\mathcal{C}_t$, all the classes corresponding to dynamic objects in $\mathcal{C}_t$ are chosen as the mixed classes $\mathcal{C}_t^m$.
The binary mask $\mathcal{M}_{M\to F}$ is calculated:
\begin{equation} \label{eq:mask-2} \small
    \mathcal{M}_{M\to F}=\left \{\begin{array}{lc}
         1, & P_d^{(p)}\in \mathcal{C}_t^m \\
         0,  & otherwise
    \end{array} \right.,
\end{equation}
where $p$ denotes a pixel in a daytime image. 
This selection is based on the fact that, apart from illumination, the main difference between daytime images and nighttime images is that dynamic objects\footnote{In our experiments, the dynamic object categories involve the category person, rider, car, truck, bus, train, motorcycle, and bicycle.}  often appear during the day and disappear at night.

With the obtained mask $\mathcal{M}_{M\to F}$, the mixed sample $I_{M\to F}$ is obtained by:
\begin{equation}
    I_{M\to F}= \mathcal{M}_{M\to F}\cdot I_d + (1-\mathcal{M}_{M\to F})\cdot I_n.
\end{equation}
Then, we propose a consistency loss $\mathcal{L}_{M\to F}$, to enforce the $I_{M\to F}^{re}$ and $I_d^{re} $ to be similar, which is defined as follow:
\begin{equation}
    \mathcal{L}_{M\to F}=||I_{M\to F}^{re}-I_d^{re}||_1.
\end{equation}
}

\subsection{Optimization Objective}
\label{sec:overall}

Besides the loss functions introduced in the Trans2Seg and Seg2Trans branches, 
we also optimize the semantic segmentation model with the labeled source samples, using the cross-entropy loss:
\begin{equation}
    \mathcal{L}_{M}=-\frac{1}{N\cdot C}\sum_{c\in C} \mathbbm{1}(Y_s)^c\log P_s^c,
\end{equation}
where $P_s^c$ is the $c$-th channel of the prediction $P_s$ and $\mathbbm{1}(\cdot)$ means the one-hot encoding operation.

In addition, we also train two discriminators $D_s$ and $D_n$ into our framework, adopting adversarial learning to align the distributions of segmentation outputs between source and target domains. Instead of using the vanilla
GAN objective, we adopt the least-squares loss function~\cite{mao2017least} for stable training.
The adversarial losses can be expressed as:
\begin{equation}
\begin{aligned}
    \mathcal{L}_{adv}&= (D_d(P_d)-1)^2 + (D_n(P_n)-1)^2,\\
    \mathcal{L}_D&= (D_d(P_S)-1)^2+(D_n(P_S)-1)^2\\
    &+ (D_d(P_d))^2+(D_n(P_n))^2,
\end{aligned}  
\end{equation}
where $\mathcal{L}_{D}$ is used to optimize the two discriminators and $\mathcal{L}_{adv}$ is exploited to optimize the semantic segmentation and the translation networks.

The whole optimization objective of our BiMix is defined as:
\begin{equation}
\begin{aligned}
    \mathcal{L}&=\mathcal{L}_{enhance}+\mathcal{L}_{M}+\mathcal{L}_D+\mathcal{L}_{adv}\\
    &+\mu_{1}\mathcal{L}_{F\to M}+\mu_{2}\mathcal{L}_{M\to F}+\mu_{3}\mathcal{L}_{ssl},
\end{aligned}  
\end{equation}
where $\mu_1$, $\mu_2$ and $\mu_3$ are hyper-parameters controlling the importance of the corresponding loss term.

\begin{table*}[!t]
\caption{Comparison with state-of-the-art methods (per-category results) on Dark Zurich-test. The best results are presented in \textbf{bold}. `D' and `R' means DeepLab-v2~\cite{chen2017deeplab} and RefineNet~\cite{lin2017refinenet}, respectively.}
\label{tab:zurich-test}
\centering
\resizebox{1\textwidth}{!}{%
\begin{tabular}{lccccccccccccccccccccc}
\toprule
Method & Backbone & \rotatebox{90}{road} & \rotatebox{90}{sidewalk} & \rotatebox{90}{building} & \rotatebox{90}{wall} & \rotatebox{90}{fence} & \rotatebox{90}{pole} & \rotatebox{90}{light} & \rotatebox{90}{sign} & \rotatebox{90}{vegetation} & \rotatebox{90}{terrain} & \rotatebox{90}{sky} & \rotatebox{90}{person} & \rotatebox{90}{rider} & \rotatebox{90}{car} & \rotatebox{90}{truck} & \rotatebox{90}{bus} & \rotatebox{90}{train} & \rotatebox{90}{motocycle} & \rotatebox{90}{bicycle} & \textbf{mIoU} \\ 
\midrule
RefineNet~\cite{lin2017refinenet} & R & 68.8 & 23.2 & 46.8& 20.8 & 12.6 & 29.8 & 30.4 & 26.9 & 43.1 & 14.3 & 0.3 & 36.9 & 49.7 & 63.6 & 6.8 & 0.2 & 24.0 &33.6 &9.3 & 28.5\\
DeepLab-v2~\cite{chen2017deeplab} & D & 79.0 & 21.8 & 53.0 & 13.3 & 11.2 & 22.5 & 20.2 & 22.1 & 43.5 & 10.4 & 18.0 & 37.4 & 33.8 & 64.1 & 6.4 & 0.0 & 52.3 & 30.4 & 7.4 & 28.8\\
\midrule
AdaptSegNet~\cite{tsai2018learning} & D & 86.1 & 44.2 & 55.1 & 22.2 & 4.8 & 21.1 & 5.6 & 16.7 & 37.2 & 8.4 & 1.2 & 35.9 & 26.7 & 68.2 & 45.1 & 0.0 & 50.1 & 33.9 & 15.6 & 30.4 \\
ADVENT~\cite{vu2019advent} & D & 85.8 & 37.9 & 55.5 & 27.7 & 14.5 & 23.1 & 14.0 & 21.1 & 32.1 & 8.7 & 2.0 & 39.9 & 16.6 & 64.0 & 13.8 & 0.0 & 58.8 & 28.5 & 20.7 & 29.7 \\
BDL~\cite{li2019bidirectional}& D & 85.3 & 41.1 & 61.9 & 32.7 & 17.4 & 20.6 & 11.4 & 21.3 & 29.4 & 8.9 & 1.1 & 37.4 & 22.1 & 63.2 & 28.2 & 0.0 & 47.7 & 39.4 & 15.7 & 30.8 \\
\midrule
DMAda~\cite{dai2018dark} & R & 75.5 & 29.1 & 48.6 & 21.3 & 14.3 & 34.3 & 36.8 & 29.9 & 49.4 & 13.8 & 0.4 & 43.3 & 50.2 & 69.4 & 18.4 & 0.0 & 27.6 & 34.9 & 11.9 & 32.1 \\
GCMA~\cite{sakaridis2019guided} & R & 81.7 & 46.9 & 58.8 & 22.0 & 20.0 & 41.2 & 40.5 & \textbf{41.6} & 64.8 & \textbf{31.0} & 32.1 & \textbf{53.5} & 47.5 & \textbf{75.5} & 39.2 & 0.0 & 49.6 & 30.7 & 21.0 & 42.0 \\
MGCDA~\cite{SDV20} & R & 80.3 & 49.3 & 66.2 & 7.8 & 11.0 & 41.4 & \textbf{38.9} & 39.0 & 64.1 & 18.0 & 55.8 & 52.1 & \textbf{53.5} & 74.7 & \textbf{66.0} & 0.0 & 37.5 & 29.1 & 22.7 & 42.5 \\
DANNet~\cite{wu2021dannet} & R & 90.0 & 54.0 & 74.8 & 41.0 & \textbf{21.1} & 25.0 & 26.8 & 30.2 & \textbf{72.0} & 26.2 & \textbf{84.0} & 47.0 & 33.9 & 68.2 & 19.0 & 0.3 & \textbf{66.4} & 38.3 & 23.6 & 44.3 \\
CDAda~\cite{Xu2021CDAda} & R & \textbf{90.5} & \textbf{60.6} & 67.9 & 37.0 & 19.3 & \textbf{42.9} & 36.4 & 35.3 & 66.9 & 24.4 & 79.8 & 45.4 & 42.9 & 70.8 & 51.7 & 0.0 & 29.7 & 27.7 & \textbf{26.2} & 45.0 \\
\midrule
Bi-Mix (Ours) & R & 89.2 & 59.4 & \textbf{75.8} & \textbf{41.7} & 19.2 & 39.0 & 31.9 & 31.5 & 70.9 & 30.1 & 81.9 & 44.9 & 41.8 & 66.3 & 34.2 & \textbf{1.0} & 61.1 & \textbf{47.4} & 14.6 & \textbf{46.5} \\ 
\bottomrule
\end{tabular}}
\end{table*}

\section{Experiments}
\subsection{Datasets}

\noindent\textbf{Cityscapes~\cite{cordts2016cityscapes}} includes 5,000 images taken in street scenes with pixel-level annotations for a total of 19 categories. The image resolution is $2,048 {\times} 1,024$.
It is split into training, validation, and testing sets, including 2,975, 500, and 1,525 images, respectively. In our DANSS task, we use the training set as the labeled source domain.

\par\noindent\textbf{Dark Zurich~\cite{sakaridis2019guided}} consists of 2,416 nighttime images, 2,920 twilight images, and 3,041 daytime images for training, which are all unlabeled with a resolution of $1,920 {\times} 1,080$. Following previous works~\cite{wu2021dannet,Xu2021CDAda}, we use 2,416 night-day image pairs as the target dataset while ignoring the twilight images. Dark Zurich also provides 201 finely annotated
nighttime images divided into the validation (Dark-Zurich val) and test part (Dark-Zurich test) with 50 images and 151 images, respectively. The ground truth of
the Dark Zurich-test is not publicly available. Our results on this dataset are obtained by submitting the prediction results to the online evaluation website. We also use the Dark Zurich validation set for method evaluation.

\noindent\textbf{Nighttime Driving~\cite{dai2018dark}}
provides 50 nighttime images with the resolution of $1,920 {\times} 1,080$, which are pixel-wise annotated by the same 19 Cityscapes category labels. In our experiments, the test set is utilized for evaluation. 

\subsection{Implementation Details}
We implement our framework in PyTorch~\cite{paszke2019pytorch}.
Following~\cite{chen2017deeplab}, we train our network using the Stochastic Gradient Descent (SGD) optimizer with a momentum of 0.9 and a weight decay of $5{\times} 10^{-4}$. The initial learning rate is set to $2.5 {\times} 10^{-4}$, and then we employ the polynomial learning rate policy to decrease it with a power of 0.9. We use the Adam optimizer~\cite{kingma2014adam} for training the discriminators with $\beta$ being set to (0.9, 0.99). Following~\cite{wu2021dannet}, the hyper-parameters $\alpha_{tv}$, $\alpha_{exp}$ and $\alpha_{ssim}$ are set to 10, 1, and 1, respectively, in all experiments.
Meanwhile, the hyper-parameters $\mu_1$, $\mu_2$ and $\mu_3$ are empirically set to  0.001, 0.001 and 1 in all our experiments, respectively. The kernel size of average pooling function $\psi$ is fixed to $32{\times} 32$ .
We set the batch size to 2 and train the model for 100K iterations.
For fair comparison, we 
we adopt the commonly used RefineNet~\cite{lin2017refinenet} as the backbone of our method.
Note that, before the adaptation process, the model is first pre-trained on the Cityscapes training set for 150K iterations.

\begin{table}[!t]
\caption{Comparison with state-of-the-art methods on Nighttime Driving test set and Dark Zurich val set. The best results are presented in \textbf{bold}. `D', `R' means DeepLab-v2~\cite{chen2017deeplab} and RefineNet~\cite{lin2017refinenet}, respectively. $*$ means the checkpoint is different with the best checkpoint of Dark Zurich-test.}
\label{tab:nightdriving-test}
\centering
\resizebox{1\linewidth}{!}{%
\begin{tabular}{lccc}
\toprule
\multirow{2}{*}{Method} & \multirow{2}{*}{Backbone} & Dark Zurich & Night Driving  \\ \cmidrule{3-4} 
& & mIoU & mIoU\\
\midrule
AdaptSegNet~\cite{tsai2018learning} & D &20.2 &34.5 \\
ADVENT~\cite{vu2019advent} & D  & - &34.7 \\
BDL~\cite{li2019bidirectional} & D  &- &34.7 \\

\midrule
DMAda~\cite{dai2018dark} & R &- &36.1 \\
DANNet~\cite{wu2021dannet} & R &30.0 &42.4 \\
GCMA~\cite{sakaridis2019guided} & R &26.7 &45.6 \\
MGCDA~\cite{SDV20} & R & 26.1 &49.4 \\
CDAda~\cite{Xu2021CDAda} & R & \textbf{36.0} &50.9 \\
\midrule
Bi-Mix (Ours) & R &33.4 &47.3 \\
Bi-Mix$^*$ (Ours) & R & 35.2 &\textbf{51.6} \\
\bottomrule
\end{tabular}}
\end{table}

\begin{figure*}[!t]
\centering
\subfloat[Input]{
    \begin{minipage}{0.16\linewidth}
        \centering
        \includegraphics[width=0.993\textwidth,height=0.5in]{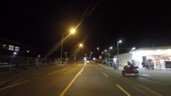}\\
        \includegraphics[width=0.993\textwidth,height=0.5in]{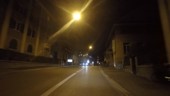}\\
        \includegraphics[width=0.993\textwidth,height=0.5in]{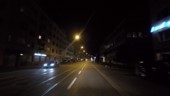}\\  
        \includegraphics[width=0.993\textwidth,height=0.5in]{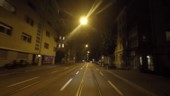}\\
    \end{minipage}%
}%
\subfloat[GCMA~\cite{sakaridis2019guided}]{
    \begin{minipage}{0.16\linewidth}
        \centering
        \includegraphics[width=0.993\textwidth,height=0.5in]{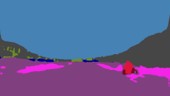}\\
        \includegraphics[width=0.993\textwidth,height=0.5in]{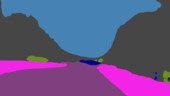}\\
        \includegraphics[width=0.993\textwidth,height=0.5in]{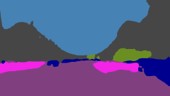}\\  
        \includegraphics[width=0.993\textwidth,height=0.5in]{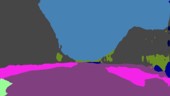}\\
    \end{minipage}%
}%
\subfloat[MGCDA~\cite{SDV20}]{
    \begin{minipage}{0.16\linewidth}
        \centering
        \includegraphics[width=0.993\textwidth,height=0.5in]{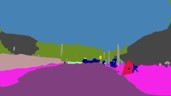}\\
        \includegraphics[width=0.993\textwidth,height=0.5in]{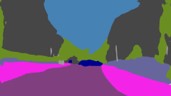}\\
        \includegraphics[width=0.993\textwidth,height=0.5in]{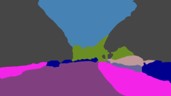}\\  
        \includegraphics[width=0.993\textwidth,height=0.5in]{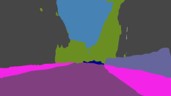}\\
    \end{minipage}%
}%
\subfloat[DANNet~\cite{wu2021dannet}]{
    \begin{minipage}{0.16\linewidth}
        \centering
        \includegraphics[width=0.993\textwidth,height=0.5in]{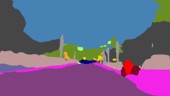}\\
        \includegraphics[width=0.993\textwidth,height=0.5in]{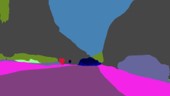}\\
        \includegraphics[width=0.993\textwidth,height=0.5in]{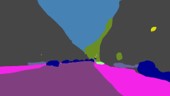}\\  
        \includegraphics[width=0.993\textwidth,height=0.5in]{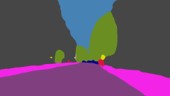}\\
    \end{minipage}%
}%
\subfloat[Bi-Mix (Ours)]{
    \begin{minipage}{0.16\linewidth}
        \centering
        \includegraphics[width=0.993\textwidth,height=0.5in]{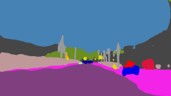}\\
        \includegraphics[width=0.993\textwidth,height=0.5in]{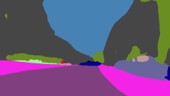}\\
        \includegraphics[width=0.993\textwidth,height=0.5in]{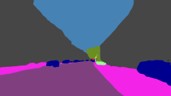}\\  
        \includegraphics[width=0.993\textwidth,height=0.5in]{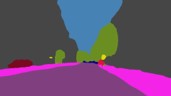}\\
    \end{minipage}%
}%
\subfloat[GT]{
    \begin{minipage}{0.16\linewidth}
        \centering
        \includegraphics[width=0.993\textwidth,height=0.5in]{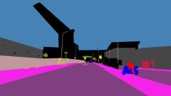}\\
        \includegraphics[width=0.993\textwidth,height=0.5in]{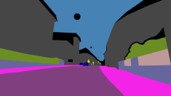}\\
        \includegraphics[width=0.993\textwidth,height=0.5in]{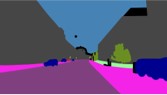}\\  
        \includegraphics[width=0.993\textwidth,height=0.5in]{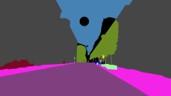}\\
    \end{minipage}%
}%
\centering
\caption{Qualitative examples on Dark Zurich-val.}
\label{fig:vis_zurich}
    \vspace{-0.4cm}
\end{figure*}

\begin{figure*}[!t]
\centering
\subfloat[Input]{
    \begin{minipage}{0.16\linewidth}
        \centering
        \includegraphics[width=0.993\textwidth,height=0.5in]{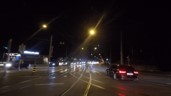}\\
        \includegraphics[width=0.993\textwidth,height=0.5in]{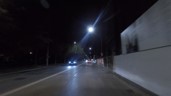}\\
        \includegraphics[width=0.993\textwidth,height=0.5in]{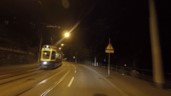}\\  
        \includegraphics[width=0.993\textwidth,height=0.5in]{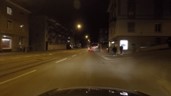}\\
    \end{minipage}%
}%
\subfloat[DANNet~\cite{wu2021dannet}]{
    \begin{minipage}{0.16\linewidth}
        \centering
        \includegraphics[width=0.993\textwidth,height=0.5in]{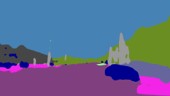}\\
        \includegraphics[width=0.993\textwidth,height=0.5in]{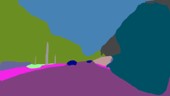}\\
        \includegraphics[width=0.993\textwidth,height=0.5in]{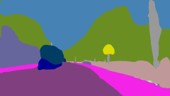}\\  
        \includegraphics[width=0.993\textwidth,height=0.5in]{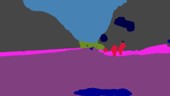}\\
    \end{minipage}%
}%
\subfloat[GCMA~\cite{sakaridis2019guided}]{
    \begin{minipage}{0.16\linewidth}
        \centering
        \includegraphics[width=0.993\textwidth,height=0.5in]{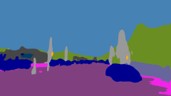}\\
        \includegraphics[width=0.993\textwidth,height=0.5in]{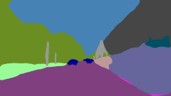}\\
        \includegraphics[width=0.993\textwidth,height=0.5in]{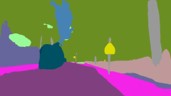}\\  
        \includegraphics[width=0.993\textwidth,height=0.5in]{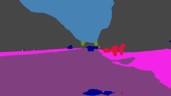}\\
    \end{minipage}%
}%
\subfloat[MGCDA~\cite{SDV20}]{
    \begin{minipage}{0.16\linewidth}
        \centering
        \includegraphics[width=0.993\textwidth,height=0.5in]{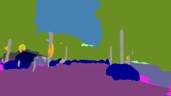}\\
        \includegraphics[width=0.993\textwidth,height=0.5in]{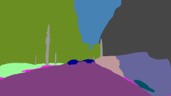}\\
        \includegraphics[width=0.993\textwidth,height=0.5in]{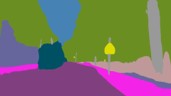}\\  
        \includegraphics[width=0.993\textwidth,height=0.5in]{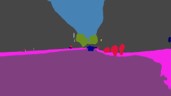}\\
    \end{minipage}%
}%
\subfloat[Bi-Mix (Ours)]{
    \begin{minipage}{0.16\linewidth}
        \centering
        \includegraphics[width=0.993\textwidth,height=0.5in]{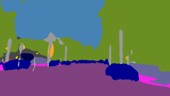}\\
        \includegraphics[width=0.993\textwidth,height=0.5in]{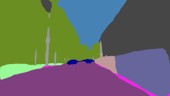}\\
        \includegraphics[width=0.993\textwidth,height=0.5in]{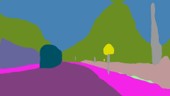}\\  
        \includegraphics[width=0.993\textwidth,height=0.5in]{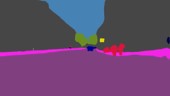}\\
    \end{minipage}%
}%
\subfloat[GT]{
    \begin{minipage}{0.16\linewidth}
        \centering
        \includegraphics[width=0.993\textwidth,height=0.5in]{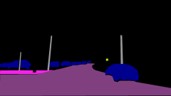}\\
        \includegraphics[width=0.993\textwidth,height=0.5in]{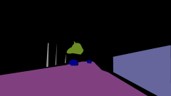}\\
        \includegraphics[width=0.993\textwidth,height=0.5in]{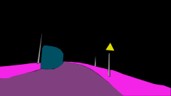}\\  
        \includegraphics[width=0.993\textwidth,height=0.5in]{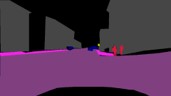}\\
    \end{minipage}%
}%
\centering
\caption{Qualitative examples on Night Driving-test.}
\vspace{-0.3cm} 
\label{fig:vis_driving}
\end{figure*}

\subsection{Comparison with State-of-the-Art Methods}
\label{sec:result}
\par\noindent\textbf{Results on Dark Zurich-test.}
We compare Bi-Mix with state-of-the-art methods for DANSS, including CDAda~\cite{Xu2021CDAda}, DANNet~\cite{wu2021dannet}, MGCDA~\cite{SDV20}, GCMA~\cite{sakaridis2019guided}, and DMAda~\cite{dai2018dark}. We also compare with some traditional domain adaptation approaches,~\ie, BDL~\cite{li2019bidirectional}, ADVENT~\cite{vu2019advent}, and AdaptSegNet~\cite{tsai2018learning}. We report the mIoU performance for all the methods in Table~\ref{tab:zurich-test}. Note that AdaptSeg, BDL, and ADVENT are trained from Cityscapes to Dark Zurich-night set.
{Considering that different baseline models are used for the previous methods, we also report the performance of the corresponding baseline Cityscapes models for the above
methods to conduct a fair comparison.}
Among these methods, RefineNet is used in DMAda, GCMA, MGCDA, DANNet, CDAda, and Bi-Mix, while the rest are based on Deeplab-v2. ResNet-101~\cite{he2016deep} is chosen as the backbone of the two baseline models, which allows us to make a direct comparison.

Our Bi-Mix achieve a \textbf{1.5\%} gain in term of mIoU over the best score obtained by the strongest existing method (CDAda). 
We also observe that our Bi-Mix significantly outperforms other methods on quite a few categories, such as building, wall, bus, and bicycle, which indicates that our method solves the large day-to-night domain gap much effectively even in low discernible regions.
In particular, compared with a similar method, BDL~\cite{li2019bidirectional}, Bi-Mix obtains a significant increase of 15.7\% in performance.
It proves that Bi-Mix is more robust and that the translation and segmentation adaptation models operate well in sinergy. 
Sample qualitative results on the Dark Zurich validation are provided in Figure~\ref{fig:vis_zurich}, confirming our quantitative results.

\par\noindent\textbf{Results on Night Driving.}
To demonstrate the generality of our approach, we also performed experiments on the Night Driving test set. Quantitative and qualitative results are shown in Table~\ref{tab:nightdriving-test} and in Figure~\ref{fig:vis_driving}, respectively. Our method is by far the best performing adaptation approach. It is worth mentioning that the Night Driving dataset is not labeled as accurately as Dark Zurich-test as shown in Figure~\ref{fig:vis_driving}. Even with these issues, our Bi-Mix achieves 0.7\% improvement of the overall mIoU over the leading methods. Bi-Mix also surpasses DANNet~\cite{wu2021dannet} by a significant margin (9.2\%). In our experiments, we also observe one phenomenon: the checkpoint when the model gets the best performance in the Nighttime Driving is different from when the model gets the best performance in the Dark Zurich-test. We report both results in Table~\ref{tab:nightdriving-test} for a fair comparison. 

\begin{figure*}[t]
\centering
\subfloat[Night]{
    \begin{minipage}{0.16\linewidth}
        \centering
        \includegraphics[width=0.993\textwidth,height=0.5in]{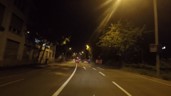}\\
        \includegraphics[width=0.993\textwidth,height=0.5in]{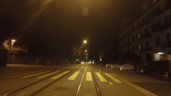}\\
    \end{minipage}%
}%
\subfloat[Baseline Relight]{
    \begin{minipage}{0.16\linewidth}
        \centering
        \includegraphics[width=0.993\textwidth,height=0.5in]{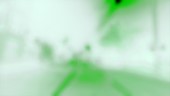}\\
        \includegraphics[width=0.993\textwidth,height=0.5in]{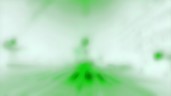}\\
    \end{minipage}%
}%
\subfloat[Baseline Enhancement]{
    \begin{minipage}{0.17\linewidth}
        \centering
        \includegraphics[width=0.993\textwidth,height=0.5in]{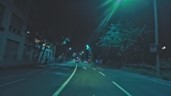}\\
        \includegraphics[width=0.993\textwidth,height=0.5in]{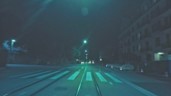}\\
    \end{minipage}%
}%
\subfloat[Bi-Mix Relight]{
    \begin{minipage}{0.16\linewidth}
        \centering
        \includegraphics[width=0.993\textwidth,height=0.5in]{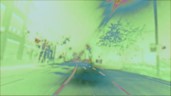}\\
        \includegraphics[width=0.993\textwidth,height=0.5in]{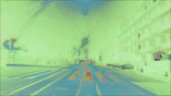}\\
    \end{minipage}%
}%
\subfloat[Bi-Mix Enhancement]{
    \begin{minipage}{0.16\linewidth}
        \centering
        \includegraphics[width=0.993\textwidth,height=0.5in]{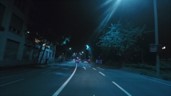}\\
        \includegraphics[width=0.993\textwidth,height=0.5in]{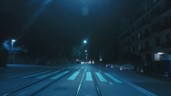}\\
    \end{minipage}%
}%
\subfloat[Day]{
    \begin{minipage}{0.16\linewidth}
        \centering
        \includegraphics[width=0.993\textwidth,height=0.5in]{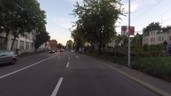}\\
        \includegraphics[width=0.993\textwidth,height=0.5in]{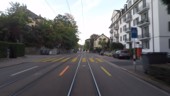}\\
    \end{minipage}%
}%
\centering
\caption{Visualization results of relighted images and enhanced images on Dark-Zurich val dataset.}
\label{fig:vis_ab}
\end{figure*}

\begin{table*}[!t]
\caption{Ablation study on Bidirectional Mixed Sampling. `F' and `M' means translation network and segmentation network, respectively.}
\label{tab:ab_dms}
\centering
\resizebox{0.9\textwidth}{!}{%
\begin{tabular}{l|cc|c|ccc}
\toprule
\multirow{2}{*}{Method} & \multicolumn{2}{c|}{Mixed Sampling} & \multirow{2}{*}{Relight Net} & {Dark-Zurich val} & {Dark-Zurich test} & {Nighttime Driving} \\ \cmidrule{2-3} \cmidrule{5-7} 
 & M$\to$F  & F$\to$M & & mIoU & mIoU & mIoU\\
\midrule
Baseline &   &   &  $\checkmark$ & 30.4 & 44.3 & 42.3\\
\midrule
\multirow{3}{*}{Unidirectional Mixed Sampling} & $\checkmark$ &   &  $\checkmark$ & 33.2 & 45.4 & 47.1 \\
 &   &  $\checkmark$ &   & 33.2 & 44.4 & 43.7\\
 &   &  $\checkmark$ &  $\checkmark$ & \textbf{34.0} & 45.5 & 46.4\\
\midrule                                            
Bidirectional Mixed Sampling &  $\checkmark$ &  $\checkmark$ &  $\checkmark$ & {33.4} & \textbf{46.5} & \textbf{47.3}  \\ 
\bottomrule
\end{tabular}}
\end{table*}

\subsection{Ablation Study}

To demonstrate the effectiveness of different components of Bi-Mix, we train several model variants and test them on Dark Zurich-val, Dark Zurich-test, and Night Driving-test. The performance results are reported in Table~\ref{tab:ab_dms}. We choose DANNet~\cite{wu2021dannet} as our baseline and the reproduced results are reported in the first row of Table~\ref{tab:ab_dms}. Bi-Mix indicates our complete method, while Unidirectional Mixed Sampling (UDMS) means that mixed sampling is used in only one direction. Unsurprisingly, UDMS on the Trans2Seg branch and Seg2Trans branch provides improvements compared to the baseline results. This also proves that while noise may appear in prediction at the beginning due to the nature of the mixing strategy on the Trans2Seg branch, its effect becomes smaller as training progresses. 
Moreover, consistency regularisation boosts performance with even imperfect labels.

We also compare UMDS with and without relighting net in Table~\ref{tab:ab_dms}. Looking at results in the third and the fourth row, it is clear that the model without relighting net produces a clear degradation in performance on the three datasets, proving the effectiveness of the relighting net. Visualization results of the relighted images and enhanced images on Dark-Zurich val dataset are shown in Figure~\ref{fig:vis_ab}. The segmentation outputs of Bi-Mix's relight net are of much higher quality than the baseline. Finally, overall, these qualitative and quantitative results confirm that in Bi-Mix the translation model and the segmentation adaptation model well complement each other.

\begin{figure}[!t]
    \centering
    \includegraphics[width=0.8\linewidth]{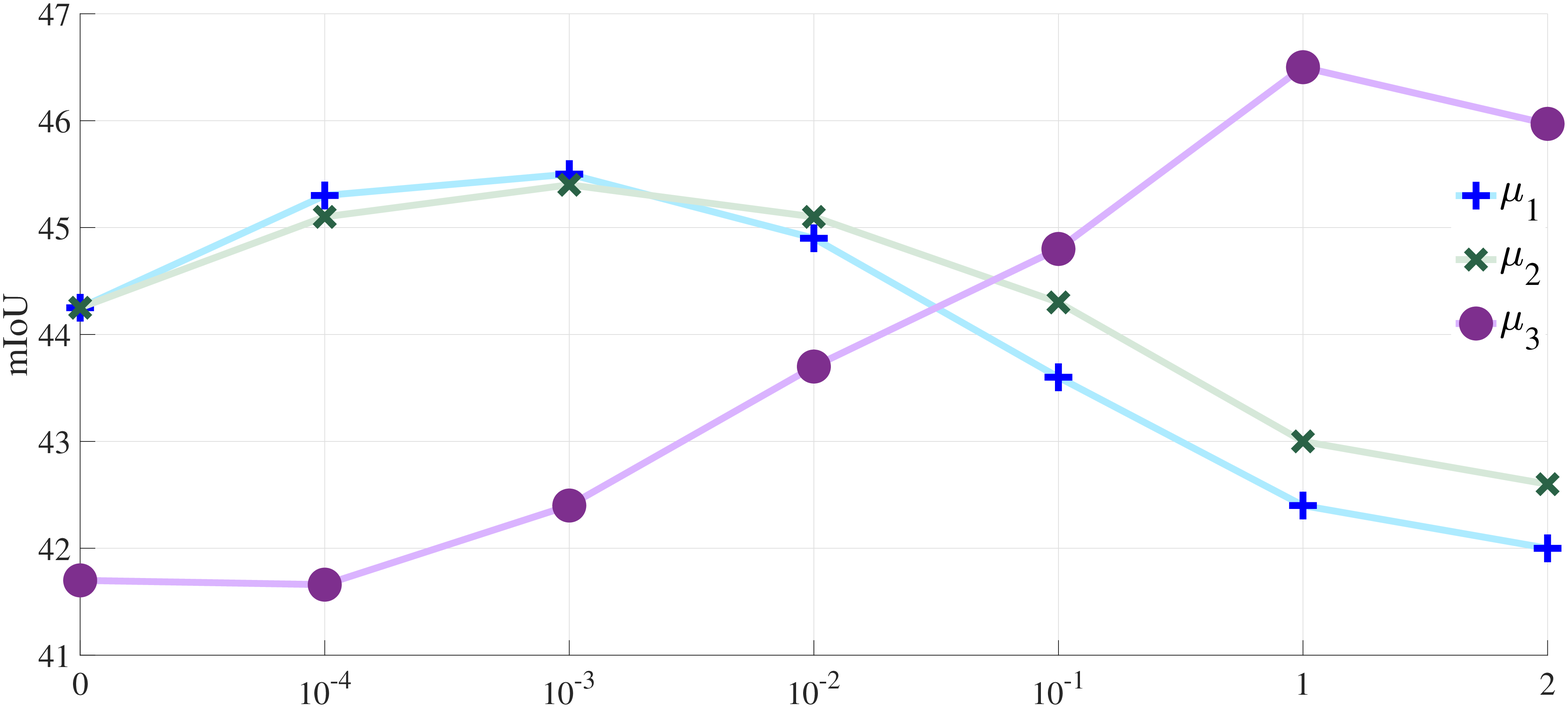}
    \caption{Performance with the different $\mu_1$, $\mu_2$, and $\mu_3$. }
    \label{fig:sensitivity}
\end{figure}

\subsection{Parameter Analysis}
We investigate the sensitivity of our method to the hyper-parameters $\mu_1$, $\mu_2$, and $\mu_3$. Specifically, when evaluating the values of $\mu_1$ or $\mu_2$, we remove the unidirectional mixed sampling on the Seg2Trans branch or Trans2Seg branch. It helps us to analyze better the impact of unidirectional mixed sampling of each branch individually. Results obtained on the Dark-Zurich test dataset are shown in Figure~\ref{fig:sensitivity}. When $\mu_1$=0 or  $\mu_2$=0, the model reduces to the baseline without using any unidirectional mixed sampling. We can find that assigning a small value (\textit{e.g.}, $<$1e-3) to $\mu_1$ and  $\mu_2$ improves the performance of the baseline while increasing the value will decrease the performance. Adding self-supervised learning can also improve the model performance, which achieves the best performance when its weight ($\mu_3$) is 1. According these results, we choose $\mu_1$, $\mu_2$, and $\mu_3$ as 0.001, 0.001, and 1, respectively.

\subsection{Limitations}
As reported in Table~\ref{tab:nightdriving-test} (last two rows), the best checkpoints are different on the Dark Zurich and Night Driving. Specifically, the best checkpoint of Dark Zurich achieves 47.3\% mIoU on Night Driving, which is lower than that of the best checkpoint selected on the Night Driving (51.6\%). The main reason is that the model is somewhat overfitting on the Dark Zurich dataset. This is a domain generalization problem and we would like to leave it to future study.
$\mu_2$ is tested by the final Bi-Mix network.

\section{Conclusions}
This paper proposes a novel bidirectional mixing framework for DANSS. Our framework involves two separated modules: an image-to-image translation model and a segmentation adaptation
model. The learning process
involves two directions. In the Trans2Seg direction, we adopt a segmentation-based data augmentation strategy, \emph{i.e.}, the mixed sampling. Similarly, we use a mixed sampling strategy to insert half of the predicted classes from daytime images into nighttime images and feed the image-to-image translation networks on the Seg2Trans direction. 
Experimental results demonstrate the effectiveness of our BiMix method which achieves state-of-the-art performance on the Dark-Zurich and the Night Driving test datasets. 

\vspace{.05in}
\noindent\textbf{Broader Impact.}
Our BiMix approach for DANSS is quite generic, and it can be applied to several settings in autonomous driving scenes.
On the bright side, this could accelerate the adoption of autonomous driving technologies to the market, improving people's safety. For instance, BiMix and other approaches for DANSS can be used to avoid collisions, detect obstacles and pedestrians at night, thus preventing car accidents. 
On the negative side, there exist malicious applications of technologies for DANSS. For instance, these technologies can facilitate activities of behavior control or people monitoring at night. However, we believe that the benefits outweigh the potential threats.

\clearpage
{\small
\bibliographystyle{ieee_fullname}
\bibliography{egbib}
}

\end{document}


\title{Supplementary Material}  
\maketitle
We also make ablation study on mixed sampling strategy to analysis the impact of mixed sampling strategy on performance. The result is shown in Table~\ref{tab:sm-mss}. Similar to the ablation study on the main paper, the Unidirectional Mixed Sampling on the Trans2Seg branch and Seg2Trans branch provides improvements, compared to the baseline results. We use randomly choosing  and choosing dynamic classes for Trans2Seg branch and Seg2Trans branch, respectively. Because  adding mixed sampling case the best checkpoints are different on the Dark Zurich and Night Driving, we choose the final mixed sampling strategy on which the result is more suitable in Table~\ref{tab:sm-mss}. In case, we use the randomly choosing mixed sampling strategy on Trans2Seg branch and choosing dynamic classes on Seg2Trans branch for our Bi-Mix.

\begin{table*}[h]
\caption{Ablation study on Mixed Sampling strategy. R and D mean randomly choosing and choosing dynamic classes .}
\label{tab:sm-mss}
\centering
\resizebox{.6\textwidth}{!}{%
\begin{tabular}{ccccc}
\toprule
\multirow{2}{*}{Trans2Seg} & \multirow{2}{*}{Seg2Trans} & Dark-Zurich val & Dark-Zurich test & Nighttime Driving \\\cmidrule{3-5} 
& & mIoU & mIoU & mIoU \\
\midrule
& & 30.4 & 44.3 & 42.3 \\
D & & 34.0 & 45.5 & 46.4 \\
R & & \textbf{35.2} & 45.4 & \textbf{50.5} \\
& D & 33.2 & 45.4 & 47.6 \\
& R & 34.2 & 44.4 & 43.7 \\
R & D & 33.4 & \textbf{46.5} & 47.3 \\   
\bottomrule
\end{tabular}}
\end{table*}
\clearpage

{\small
\bibliographystyle{ieee_fullname}
\bibliography{egbib}
}